# RANKING XAI METHODS FOR HEAD AND NECK CANCER OUTCOME PREDICTION

*Baoqiang Ma, Djennifer K. Madzia-Madzou, Rosa C.J. Kraaijveld, Jin Ouyang*

Image Sciences Institute, University Medical Center Utrecht

## ABSTRACT

For head and neck cancer (HNC) patients, prognostic outcome prediction can support personalized treatment strategy selection. Improving prediction performance of HNC outcomes has been extensively explored by using advanced artificial intelligence (AI) techniques on PET/CT data. However, the interpretability of AI remains a critical obstacle for its clinical adoption. Unlike previous HNC studies that empirically selected explainable AI (XAI) techniques, we are the first to comprehensively evaluate and rank **13 XAI methods across 24 metrics**, covering **faithfulness, robustness, complexity and plausibility**. Experimental results on the multi-center HECKTOR challenge dataset show large variations across evaluation aspects among different XAI methods, with Integrated **Gradients (IG) and DeepLIFT (DL)** consistently obtained high rankings for faithfulness, complexity and plausibility. This work highlights the importance of comprehensive XAI method evaluation and can be extended to other medical imaging tasks.

## 1. INTRODUCTION

Head and neck cancer (HNC) is the seventh most common cancer worldwide [1], treated primarily with radiotherapy with or without chemotherapy and surgery. Despite similar treatments, substantial variability in outcomes remains among patients. This motivates the development of predictive models to guide personalized treatment.

Recent studies have applied deep learning techniques for PET/CT-based HNC outcome prediction, surpassing traditional radiomics approaches. The HECKTOR challenge [2], [3] is a prominent benchmark providing a multi-center PET/CT dataset and fostering competition for improved prediction performance. State-of-the-art (SOTA) methods typically employ CNN- and/or transformer-based architectures such as TransRP [4], [5] and XSurv [6]. Interestingly, recent work has shown that DenseNet-based models achieve comparable or even superior performance on external test sets [7].

Compared with the extensive investigation of network architecture, the explainability of these models has been much less explored, despite being essential for clinical applications. For instance, TransRP and XSurv employed Grad-CAM and attention rollout to generate coarse saliency maps for visual plausibility (i.e., alignment with human intuition). However, they neither quantitatively evaluated these explanations nor considered other aspects such as faithfulness (i.e., how well explanations reflect true model reasoning). As a result, their selected XAI methods remain less reliable, highlighting the need for a more systematic and objective evaluation of explainability methods.

Inspired by the LATEC benchmark [8], which systematically evaluated and ranked 17 XAI methods across 20 metrics for faithfulness, robustness and complexity in different image tasks, we aim to extend this idea to the specific task of HNC outcome prediction. The LATEC study revealed large variations among XAI approaches across metrics, indicating that no single approach was best overall. Therefore, building on this insight, we **present the first comprehensive evaluation and ranking of 13 XAI methods** for HNC outcome prediction on the HECKTOR dataset, covering **24 metrics across faithfulness, robustness, complexity and plausibility**—an essential and clinically relevant aspect not addressed in LATEC.

## 2. MATERIALS AND METHODS

### 2.1. Dataset

The latest HECKTOR 2025 training dataset (https://hecktor25.grand-challenge.org/dataset/) was used to develop HNC outcome prediction models. Data from 651 patients, each with CT, PET and Gross Tumor Volume (GTV) mask (Fig. 1.1) of primary tumor and lymph nodes were included. The data was randomly split in a train set of 488 patients (75%) and a test set of 163 patients (25%). The predicted endpoint of the model was recurrence-free survival (RFS). All images were center cropped at the GTV with a size of 192×192×192 mm³ and resampled to 2×2×2 mm³. CT and PET intensities were truncated to ranges of [-200, 200] and [0, 25], respectively, and normalized to [0, 1]. The mean values of CT, PET and GTV volumes were used as the model input.

### 2.2. Outcome prediction model

We employed a 3D DenseNet121 architecture for outcome prediction, due to its simplicity and strong performance demonstrated in previous HNC study [7] and challenges.

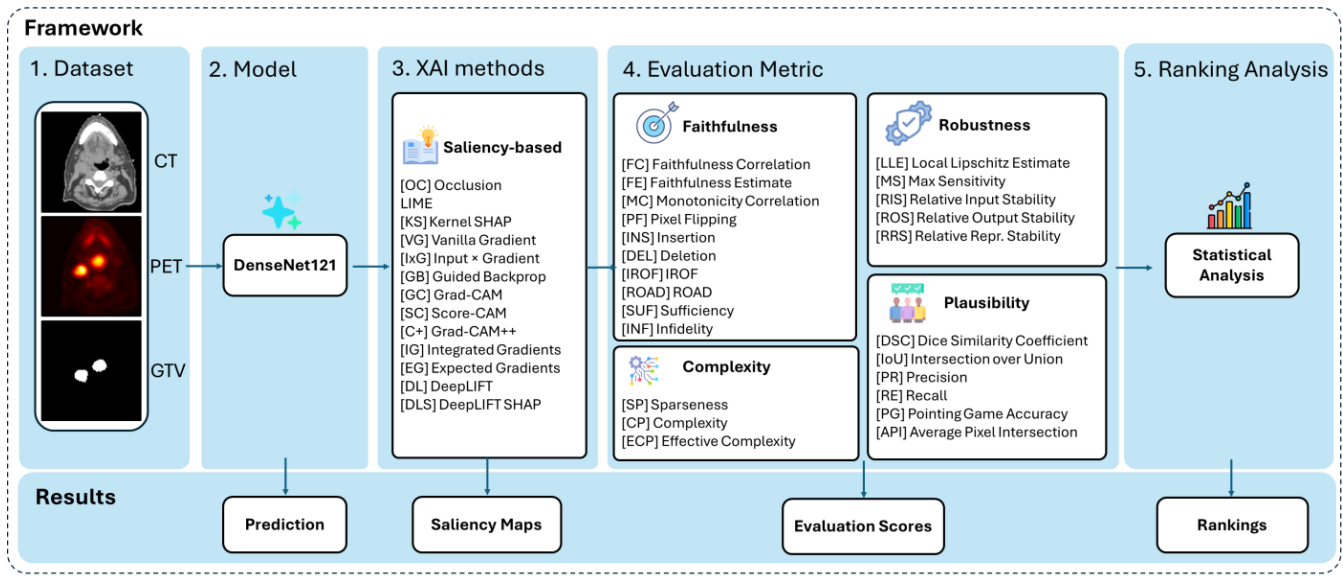

**Fig. 1.** The framework of ranking XAI methods on HNC outcome prediction

The model outputs a continuous risk score for time-to-event prediction and was trained using the Cox negative log partial likelihood loss [9].

### 2.3. XAI methods
A total of 13 saliency-based XAI methods were applied to generate attribution maps for patients in the test set, as illustrated in Fig. 1.3. These methods can be grouped into three categories: (1) perturbation-based, which evaluate feature importance by masking or altering image regions and measuring output change (e.g., OC and LIME); (2) gradient-based, which compute input attributions via backpropagated gradients (e.g., VG, IG and DL); and (3) CAM-based, which use activation maps and gradients from convolutional layers to localize predictive regions (e.g., GC, SC and C+).

### 2.4. XAI evaluation metrics
A total of 24 evaluation metrics (Fig. 1.4.) were employed to assess the quality of the generated saliency maps, covering four aspects: faithfulness (agreement with true model reasoning, 10 metrics); robustness (stability under small perturbations or noises, 5 metrics); complexity (conciseness and sparsity of highlighted regions, 3 metrics); and plausibility (alignment with clinically relevant anatomy, 6 metrics).

### 2.5. Ranking analysis
Inspired by the LATEC benchmark, we used a ranking-based evaluation to compare different XAI methods more fairly. The LATEC benchmark argued that using rankings is better than raw metric values, because each metric has its own scale and definition, making it difficult to aggregate. Specifically, for each metric, we ranked all XAI methods according to their mean performance in the test set. We then computed the mean, median, and standard deviation of these rankings within each evaluation aspect—faithfulness, robustness, complexity, and plausibility. XAI methods achieving consistently high rankings in both faithfulness and plausibility were considered clinically preferable.

### 2.6. Implementation details
The DenseNet121 training was implemented using PyTorch 2.6.0 and MONAI 1.5.0 libraries on an NVIDIA A100 GPU (40 GB), with the default DenseNet121 networks parameters settings in MONAI. The AdamW optimizer was used with an initial learning rate of 2e-4 and weight decay of 5e-5 to train the model for 50 epochs. Data augmentation included random flipping, affine transformations (translation, rotation, and scaling), and 3D elastic deformation to increase data variability. An oversampling strategy was applied to balance event and non-event cases in the training set.

XAI methods and evaluation metrics except for plausibility metrics were implemented using the Captum and Quantus libraries, with custom extensions for 3D input support by LATEC. Evaluation of saliency maps was performed on a GPU when memory was sufficient; otherwise, computation automatically switched to CPU for memory-intensive metrics. The most time-consuming evaluation metrics were those for faithfulness and robustness, as these metrics typically require multiple forward or backward model passes per sample. Detailed hyperparameter configurations for all XAI methods and evaluation metrics will be public in https://github.com/baoqiangma96/TransRP.

## 3. RESULTS

The DenseNet121 achieved a C-index of 0.66 in the multi-center test set, which is comparable with results in previous studies [4], [7]. Tab. 1 summarizes the mean, median, and standard deviation (std) of the rankings of all XAI methods across the four evaluation aspects in the test set. In general, the ranking variances across methods are reasonably clear, with SC, IG, and DL achieving the top three positions for faithfulness, while EG, VG, and GC perform best in robustness. IxG, IG and DL rank among top 3 in both complexity and plausibility.

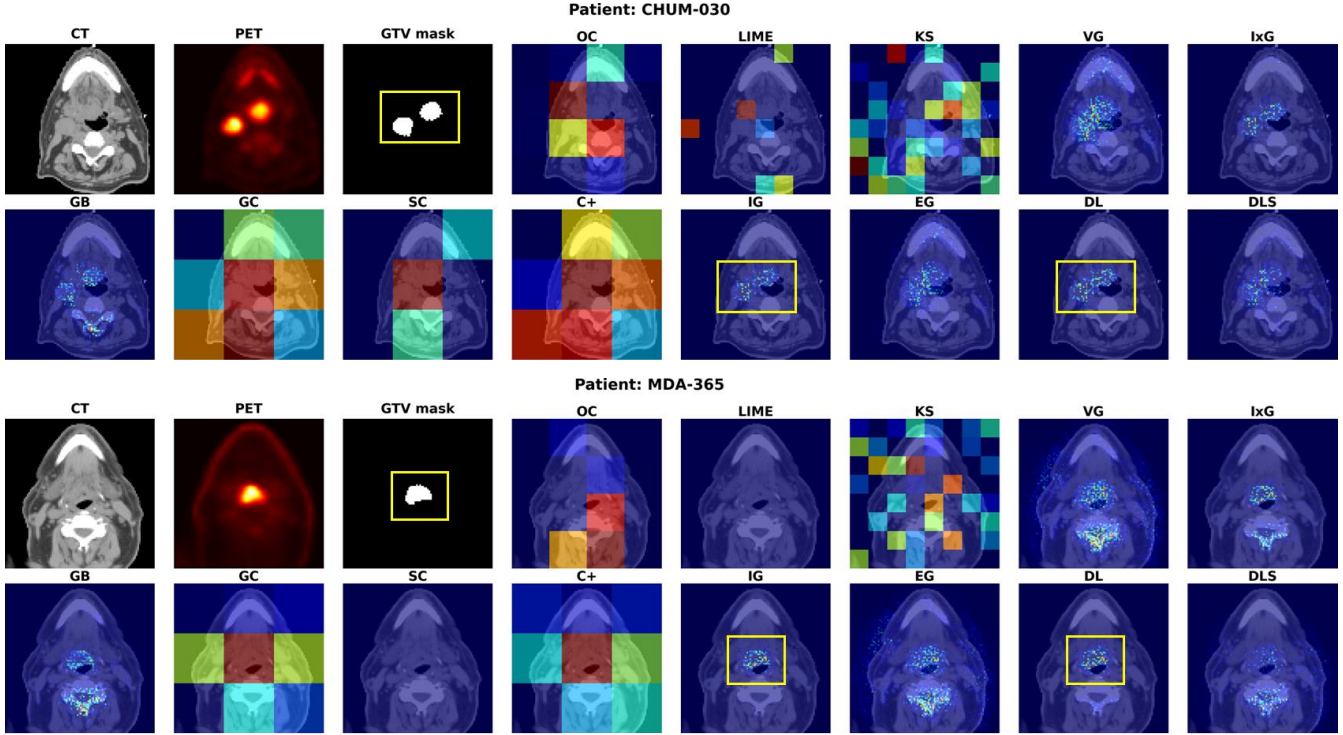

**Fig. 2.** The examples of saliency maps generated by different XAI methods.

**Tab. 1.** The rankings (mean, median and std) of each XAI method for each evaluation aspects. Orange: top 3; yellow: achieving top 3 in faithfulness, complexity and plausibility.

| Methods | Faithfulness | | | Robustness | | | Complexity | | | Plausibility | | |
|---|---|---|---|---|---|---|---|---|---|---|---|---|
| Ranking | Mean | Median | Std | Mean | Median | Std | Mean | Median | Std | Mean | Median | Std |
| OC | 6.1 | 6.0 | 5.0 | 6.2 | 6.0 | 1.1 | 5.0 | 5.0 | 0.0 | 9.7 | 10 | 0.9 |
| LIME | 5.7 | 5.5 | 4.1 | 7.2 | 7.0 | 4.3 | 9.3 | 11.0 | 2.9 | 11.2 | 11.5 | 0.8 |
| KS | 5.6 | 5.0 | 4.3 | 8.6 | 10 | 3.1 | 10.3 | 10.0 | 0.6 | 10.3 | 13 | 4.4 |
| VG | 8.4 | 10 | 4.4 | 4.2 | 2.0 | 3.2 | 7.7 | 8.0 | 1.5 | 4.7 | 4.5 | 1.1 |
| IxG | 5.8 | 5.0 | 4.5 | 10.2 | 12.0 | 4.1 | 2.0 | 2.0 | 0.0 | 3.5 | 2.5 | 2.5 |
| GB | 5.0 | 4.5 | 3.5 | 7.4 | 7.0 | 0.5 | 4.0 | 4.0 | 0.0 | 5.4 | 4.0 | 3.0 |
| GC | 5.7 | 5.0 | 4.0 | 5.0 | 4.0 | 2.3 | 12.7 | 13.0 | 0.6 | 8.2 | 8.0 | 2.7 |
| SC | 4.5 | 3.0 | 3.8 | 8.2 | 9.0 | 3.7 | 7.3 | 7.0 | 1.5 | 10.3 | 10 | 1.8 |
| C+ | 5.5 | 6.0 | 3.8 | 5.8 | 5.0 | 2.9 | 12.3 | 12.0 | 0.6 | 9.5 | 11.2 | 3.5 |
| IG | 4.7 | 3.5 | 4.1 | 10.4 | 12.0 | 4.7 | 3.0 | 3.0 | 0.0 | 2.2 | 1.0 | 2.6 |
| EG | 7.7 | 9.0 | 4.1 | 3.2 | 1.0 | 3.2 | 7.7 | 7.0 | 1.2 | 5.6 | 5.5 | 0.8 |
| DL | 4.7 | 4.0 | 3.5 | 9.4 | 11.0 | 4.8 | 1.0 | 1.0 | 0.0 | 4.0 | 4.5 | 1.0 |
| DLS | 7.2 | 9.0 | 5.0 | 5.2 | 4.0 | 2.9 | 8.7 | 8.0 | 1.2 | 6.5 | 7.0 | 1.1 |

Overall, IG and DL stand out as the best performing methods, consistently ranking within the top 3 for faithfulness, complexity, and plausibility. As displayed in Fig.2, the saliency maps generated by IG and DL show the best spatial alignment with GTV mask, indicating high plausibility. Other methods such as VG, IxG, GB, EG, and DLS also highlight tumor regions but often include other areas, such as bone structures beneath the tumor as seen in patient MDA-365 in Fig. 2. In contrast, other perturbation-based methods (OC, LIME and KS) cannot consistently localize the tumor. CAM-based approaches (GC, SC and C+) produced more global-range maps.

## 4. DISCUSSION

This study presented a comprehensive evaluation of 13 post-hoc XAI methods using 20 metrics for HNC outcome prediction task. The large standard deviations of rankings in Tab. 1 reveal substantial variations among XAI methods across metrics, which aligns with the observations from LATEC benchmark [8]. This highlights the importance of selecting XAI methods specific to a given model and dataset. While LATEC concluded that EG achieved the best overall balance of faithfulness, robustness and complexity in their 2D, 3D and point cloud datasets, we found that it performed well mainly in robustness in our dataset.

A major contribution of this work is the introduction of a comprehensive plausibility evaluation using six metrics, which are highly relevant for clinical interpretability by radiologists. Even though prior studies have suggested a plausibility does not guarantee good faithfulness [10], which was evident by VG ranking high in plausibility (4.7) but low in faithfulness (8.4) in our result, IG and DL achieved high rankings in both faithfulness and plausibility (Tab. 1). Both methods attribute predictions relative to a baseline, enabling them to trace how input features truly influence the model output (faithfulness). Moreover, their highlighted regions align well with tumor areas, likely due to the strong feature-localization ability of the DenseNet121 model. However, their dependence on a single fixed baseline and gradient propagation makes them sensitive to noise and input perturbations, explaining their lower robustness. Further work could explore how sensitive the metric rankings are to different baseline selections.

CAM-based methods (GC, SC, and C+) tended to produce more global and diffused saliency maps but often including non-tumor regions. Perturbation-based approaches (OC, LIME, KS) showed wrong or incomplete tumor localization, likely due to their sensitivity to sampling noise, occlusion patterns, and hyperparameter settings such as mask size and step size—a limitation shared by most XAI methods. Another limitation is the long runtime of faithfulness and robustness metrics, which may be improved by multi-threaded computation. Future work should explore adaptive hyperparameter optimization, metric correlation analysis and how XAI methods ranking could vary in other architectures like transformer. More importantly, human-in-the-loop evaluation with clinical experts is beneficial to confirm that quantitative metrics are indeed clinically meaningful.

## 5. CONCLUSION

In summary, this study provides a comprehensive evaluation of XAI methods for HNC outcome prediction across four aspects: faithfulness, robustness, complexity, and clinical plausibility. Integrated Gradients and Deep LIFT produced the most faithful and plausible explanations. The results underscore the need for task-specific evaluation and adaptive optimization of XAI. Future work should include clinical validation to ensure trustworthy, interpretable AI for radiotherapy applications.

## 6. COMPLIANCE WITH ETHICAL STANDARDS

## 7. ACKNOWLEDGMENTS

No funding was received for conducting this study. The authors have no relevant financial or non-financial interests to disclose. We acknowledge the idea discussion provided by Dr. Kennth Gilhuijs.